%% file: root.tex
\documentclass[letterpaper, 10 pt, conference]{ieeeconf}  

\IEEEoverridecommandlockouts                              

\usepackage{todonotes}
\usepackage{textcomp}
\usepackage{algorithm}
\usepackage[noend]{algpseudocode}
\usepackage{booktabs}
\usepackage[detect-all]{siunitx}
\usepackage{cleveref}
\usepackage[protrusion=true,expansion=true]{microtype} 
\usepackage{url}
\usepackage[keeplastbox]{flushend}

\usepackage{pgfplots}
\pgfplotsset{compat=newest}
\pgfplotsset{every axis legend/.append style={%
cells={anchor=west}}
}
\usepgfplotslibrary{polar}
\usetikzlibrary{arrows}
\tikzset{>=stealth'}
\usepgfplotslibrary{groupplots}

\usepackage[backend=bibtex,style=ieee]{biblatex}
\AtEveryBibitem{
\clearfield{url}  
\clearfield{doi}  
\ifentrytype{inproceedings}{
 	\clearlist{address}
 	\clearlist{publisher}
 	\clearname{editor}
 	\clearlist{organization}
 	\clearfield{pages}  
 	\clearlist{location}
 }{}
 }

\AtBeginBibliography{\small}

\title{\LARGE \bf
Pedestrian Collision Avoidance System for Scenarios with Occlusions
}

\author{Markus Schratter,$^{1}$ Maxime Bouton,$^{2}$ Mykel J. Kochenderfer,$^{2}$ Daniel Watzenig$^{1}$
\thanks{$^{1}$Markus Schratter and Daniel Watzenig are with the Virtual Vehicle Research Center, Graz 8010, Austria, {\tt \{markus.schratter,daniel.watzenig\}@v2c2.at}. Daniel Watzenig is also with Institute of Automation and Control, Graz University of Technology, Graz 8010, Austria }%
\thanks{$^{2}$Maxime Bouton and Mykel J. Kochenderfer are with the Department of Aeronautics and Astronautics, Stanford University, Stanford CA 94305, USA,
        {\tt \{boutonm,mykel\}@stanford.edu}.}%
}

\begin{document}

\maketitle

\thispagestyle{empty}
\pagestyle{empty}

\begin{abstract}
Safe autonomous driving in urban areas requires robust algorithms to avoid collisions with other traffic participants with limited perception ability. 
Current deployed approaches relying on Autonomous Emergency Braking (AEB) systems are often overly conservative. 
In this work, we formulate the problem as a partially observable Markov decision process (POMDP), to derive a policy robust to uncertainty in the pedestrian location. We investigate how to integrate such a policy with an AEB system that operates only when a collision is unavoidable.
In addition, we propose a rigorous evaluation methodology on a set of well defined scenarios. We show that combining the two approaches provides a robust autonomous braking system that reduces unnecessary braking caused by using the AEB system on its own. 
\end{abstract}


\section{INTRODUCTION}

Autonomous vehicles must navigate safely through urban environments where parked cars and other physical obstacles occlude other road users. 
In this work, we focus on avoiding collisions with pedestrians crossing behind an occluded area on the side of the road. 
Some systems rely on autonomous emergency braking (AEB) systems to prevent collision. They attempt to predict the trajectory of the pedestrian and compare a metric such as the time to collision (TTC) to decide when to brake~\cite{volz2018}. Although comparing TTC to a threshold to trigger braking can be effective~\cite{minderhoud2001}, it tends to be overly conservative because of the uncertainty in the sensors and environment. There is a high risk of starting unnecessary strong braking.

To provide robustness to uncertainty in the sensors and environment, previous work proposed modeling similar scenarios with occluded cars and pedestrians as partially observable Markov decision processes (POMDPs)~\cite{bouton2018, thornton2018, brechtel2014, volz2018}. Their experiments showed that POMDPs provide an effective framework for modeling uncertainty in the sensors and environment, but they assumed a different acceleration range than AEB systems. The resulting POMDP policies were designed for comfortable maneuvers and would not be able to deliver extreme deceleration when needed. Other techniques to handle planning in occluded areas rely on set based approaches~\cite{koschi2018, orzechowski2018, magdici2016}. Such methods are often well suited to achieve robust prediction and compute a safe driving velocity. However they do not offer a principle framework for combining planning and partial observability.





This paper demonstrates the benefit of augmenting a POMDP policy with an AEB system that can use the full braking power of the vehicle. We present an approach where the problem is formulated as a POMDP to derive a policy robust to uncertainty in the pedestrian state and to handle hidden pedestrian behind an occlusion. The POMDP planner is designed for comfortable maneuvers in a middle acceleration range and is responsible for taking into account uncertainty due to occlusions. The policy adapts the velocity of the vehicle when the side of the road is occluded. To handle rare critical situations where a pedestrian appears behind an occluded area while the vehicle is at high speed, an AEB system intervenes with a strong brake intervention when a collision is unavoidable. The AEB system is responsible for strong interventions when a collision is unavoidable. In situations with poor visibility, the AEB system is not able to avoid or mitigate collisions on its own. The POMDP planner enables the system to anticipate this uncertainty and to prevent an emergency stop. By combining the two systems, our algorithm is able to maintain a reasonable driving speed in occluded areas without increasing the accident rate compared to relying on the AEB system on its own. Safety is not compromised because the AEB system can take control at any time.

\begin{figure}[!t]
\centering
\includegraphics[width=0.8\columnwidth]{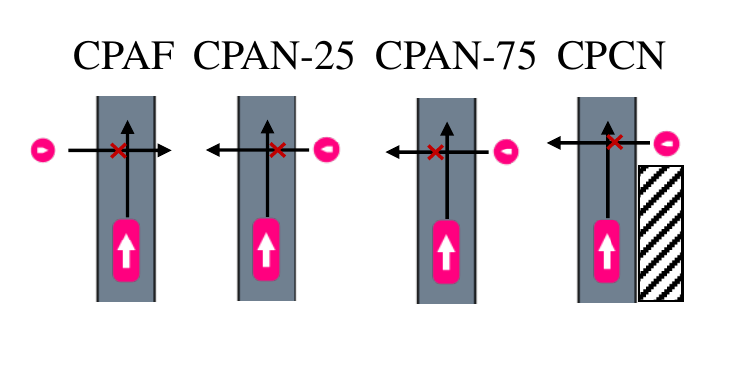}
\caption{Overview of the EuroNCAP scenarios. CPAF: Car-to-Pedestrian Farside Adult, CPAN-25/75: Car-to-Pedestrian Nearside Adult, CPCN: Car-to-Pedestrian Nearside Child with occlusion.}
\label{fig:euroncap_scenarios} 
\end{figure}

Finally, we propose a rigorous evaluation methodology on a set of well defined scenarios from the EuroNCAP test protocol (\cref{fig:euroncap_scenarios}). Previous work on evaluating autonomous braking systems at unsignalized crosswalks relied on data-driven models of the pedestrian~\cite{chen2017}. \Citeauthor{chen2017} argue that the EuroNCAP scenarios would allow an overly conservative system to be validated \cite{chen2017}. To avoid such an issue, we augmented the suite of scenarios with a situation involving occluded areas at the side of the road with no pedestrian crossing. It is expected that an efficient system would drive at a reasonable speed in these situations.

\section{Problem Formulation}

This section outlines how to model our problem as a partially observable Markov decision process and solve for an approximately optimal solution.

\subsection{Background}

Sequential decision making problems under uncertainty can be modeled as partially observable Markov decision processes (POMDPs). It is a mathematical framework defined by the tuple $(\mathcal{S}, \mathcal{A}, \mathcal{O}, T, O, R, \gamma)$ where $\mathcal{S}$ is a state space, $\mathcal{A}$ an action space, $\mathcal{O}$ an observation space, $T$ a transition model, $O$ an observation model, $R$ a reward function, and $\gamma$ a discount factor. From a state $s\in\mathcal{S}$, the agent takes an action $a$ and the state evolves to a state $s'$ with probability $T(s',s,a) = \Pr(s'\mid s, a)$. 
In a POMDP, the agent has uncertain knowledge about the state of the environment. Therefore, the agent maintains a belief about its internal knowledge of the state. The belief $b$ can be updated after taking an action $a$ and an observation $o$ about the current state using the following equation, where $T(s'\mid a, s)$ is the transition function:
\begin{equation}
b'(s')\propto O(o\mid s, a)\sum_{s}T(s'\mid a, s)b(s)
\end{equation}
In this work we used a discrete Bayesian updater, which updates the discretized belief with a measured continuous observation.

The solution of a POMDP is an optimal policy~$\pi^{*}$, which maximizes the expected discounted sum of immediate rewards from any given belief. The optimal policy can be extracted from the optimal utility function $U^{*}(b, a)$. In general, computing the exact optimal utility function for a POMDP is intractable and must rely on approximation techniques instead. Two approaches are ued to compute the optimal utility function: offline and online methods~\cite{koch2015}. In this paper, we use an offline QMDP~\cite{littman1995} approach to compute the optimal policy. The QMDP method solves the problem under the assumption that the state becomes fully observable after one time step. With this assumption the value iteration algorithm can solve the optimal state-action utility function $U^{*}(s, a)$ assuming full observability.

\subsection{Scenario modeling}

The road is represented in the Frenet frame. By applying an appropriate coordinate transform, our planner can be applied directly to different road configurations~\cite{werl2010}. For simplicity, we illustrate our approach on a straight road segment.

\subsubsection{Action space}
The POMDP planner is able to control the acceleration profile in the longitudinal direction and can position the vehicle inside the driving lane in the lateral direction along the given path. In the lateral direction, the planner can modify the vehicle position inside the lane. A finite set of strategic maneuvers for the lateral control are defined: no acceleration or an acceleration to the left or right side: ${\lbrace \SI{0}{\meter\per\second\squared}, \SI{1}{\meter\per\second\squared}, \SI{-1}{\meter\per\second\squared}\rbrace}$. Strategic maneuvers for the longitudinal control such as accelerating, maintaining constant speed and braking with different strengths are represented by a finite set of acceleration and deceleration actions:  ${\lbrace \SI{1}{\meter\per\second\squared}, \SI{0}{\meter\per\second\squared}, \SI{-1}{\meter\per\second\squared},\SI{-2}{\meter\per\second\squared}, \SI{-4}{\meter\per\second\squared} \rbrace}$. 

\subsubsection{State space}
\begin{figure}[!t]
\centering
\includegraphics[width=8cm,keepaspectratio=true]{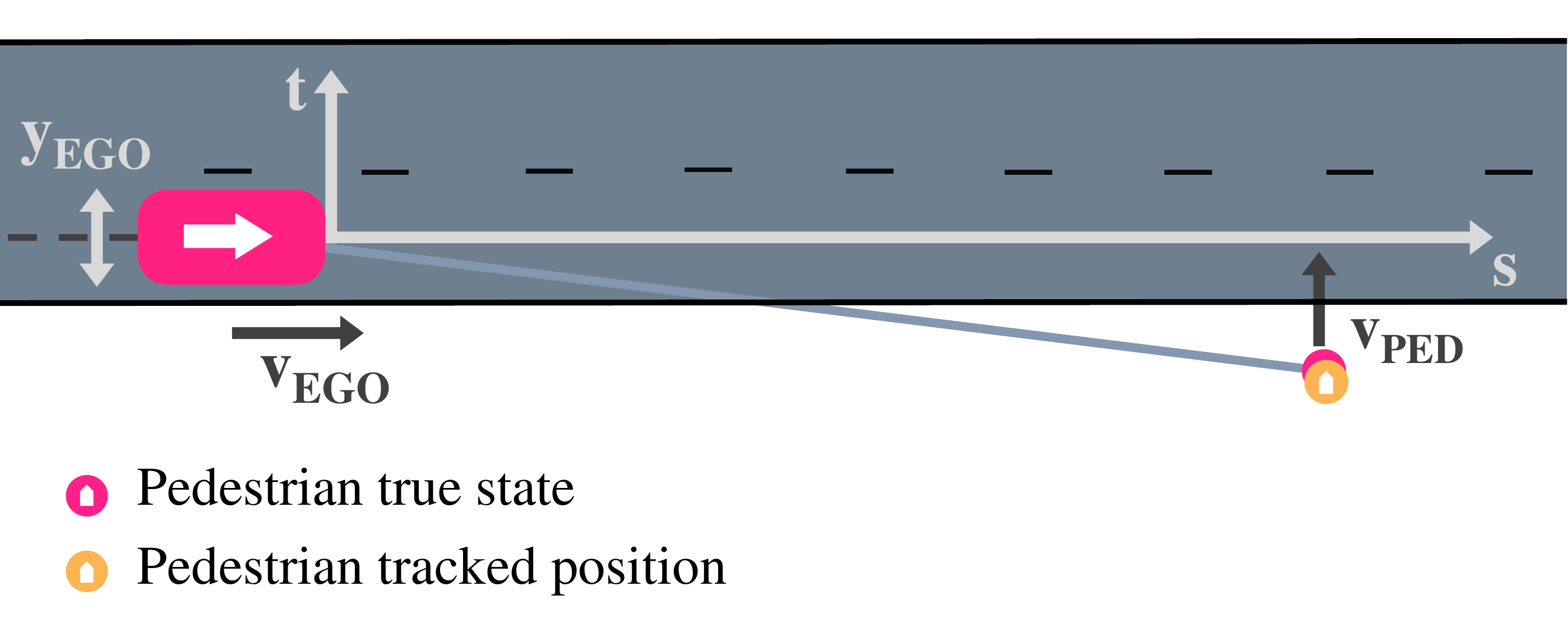}
\caption{Representation of the state of the pedestrian avoidance POMDP. The state space represents the area in front of the vehicle of a length of \SI{50}{m} and a width of \SI{10}{m}.}
\label{fig:state_space} 
\end{figure}

The state space represents all the variables taken into account for solving the problem. It encodes information on the ego vehicle and the pedestrian. To handle complex street courses the road is represented in the Frenet frame. The ego vehicle is represented in the state space with its longitudinal velocity (\SIrange{0}{50}{\kilo\meter\per\hour}) and its lateral position inside the lane ($\pm \SI{1}{\meter}$). The position of the pedestrian is represented in the longitudinal direction $s$~(\SIrange{0}{50}{\meter}) and lateral direction $t$~($\pm\SI{5}{m})$. The longitudinal range is the result of the required distance to stop based on the defined maximum velocity, the maximum deceleration of the system and a longitudinal safety gap. In addition, we consider the velocity (\SIrange{0}{7.2}{\kilo\meter\per\hour}) and orientation ($\pm\ang{90}$) of the pedestrian. \Cref{fig:state_space} illustrates the state representation with one crossing pedestrian from the left side as an example. All the variables in the state space are discretized and result in \num{29} velocities for the ego vehicle and five positions in the lane. The representation of the pedestrian needs \num{27} longitudinal positions, eleven lateral positions, five velocity levels and seven possible orientations. By multiplying all possible combinations of ego and pedestrian states, the total number of states amounts to \num{1.5e6}. 

\subsubsection{Transition model}
The transition model of the ego vehicle depends on the current action and state of the ego vehicle and consists of a point mass model. For the transition of the pedestrian, we use a simple reachability model~\cite{liur2017}, which depends on the current pedestrian state and calculates further positions for the pedestrian based on a set of possible acceleration values. It is assumes that the pedestrian can take any of those acceleration with uniform probability. The velocity of the pedestrian is bounded up to \SI{2}{\meter\per\second}. 

\subsubsection{Observation model}
The observation model characterizes what the ego vehicle can sense about the state space. We can reasonably assume that the position and velocity of the ego vehicle are nearly perfectly observable. The observation space is similar to the state space. The observation model can be described as follows:
\begin{itemize}
\item An object in a non-occluded area will always be detected.
\item An occluded object behind an obstacle will not be detected.
\item If an object is detected, the measured quantities like the position, the velocity and the orientation of the pedestrian follow a normal distribution around the true state. The parameters of the distribution depends on the perception system model.
\end{itemize} 

\subsubsection{Reward model}
The reward model defines the objective of the POMDP planner. The ego vehicle receives a penalty for colliding with a pedestrian. We define an additional reward signal to keep the velocity sufficiently high and stay in the center of the lane. If the ego vehicle drives with the desired velocity and stays in the center of the lane, no reward is received. A penalty term decreases linearly with the velocity difference and the lateral offset to the lane center. Longitudinal and lateral actions cause a penalty to avoid too many interventions. The resulting behavior of the POMDP planner can be modified by choosing different values for penalties and rewards. The values for those penalties and rewards can be tuned through simulation on defined scenarios to balance avoiding collisions and efficiency, as described in \cref{sec:Experiments}.

\subsection{Solving the optimal policy for multiple road users}

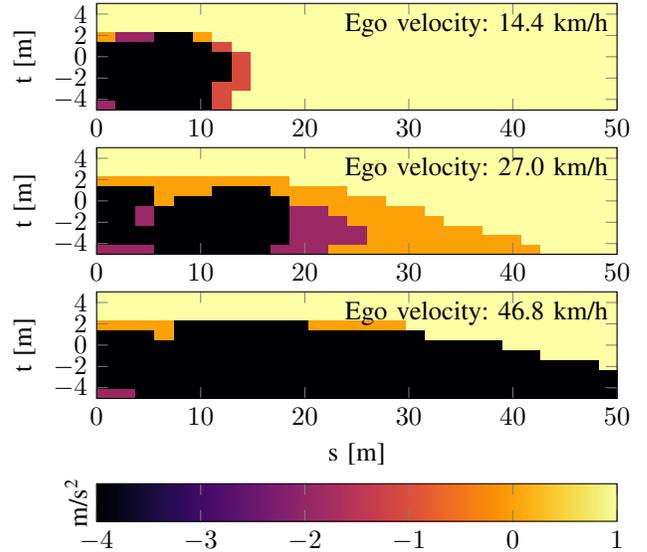
\begin{figure}[!t]
\centering
\input{policy/policyplots.tex}
\caption{Representation of different policies in variation of the ego vehicle velocity. The lateral position for the ego vehicle is fixed to the center of the driving lane. The pedestrian walks at 7.2~km/h and has an orientation of 90\textdegree. The color shows the action taken by the planner at a given longitudinal distance~[$s$] and lateral distance~[$t$].}
\label{fig:policy} 
\end{figure}

The POMDP model describds in the previous section handles only one pedestrian. To extend the capabilities of the resulting policy we use a utility decomposition method~\cite{rosenblatt1997}.
Every pedestrian is considered independently and the global utility function is approximated as the minimum belief action utility over each individual pedestrian. 
\begin{equation}
    U^{*}( b, a ) = \min_i  U_{\text{single}}^{*}( b_{i},a )
\end{equation}
where $U_{\text{single}}$ is the utility function obtained from solving the POMDP considering a single pedestrian. 
Taking the minimum will result in taking the action $a$ associated to the most critical pedestrian.  
\Cref{fig:policy} shows an example of a policy obtained by solving the POMDP model. The color shows the longitudinal action given by the policy for a given longitudinal and lateral distance in the Frenet frame $[s,\,t]$. Because the state space is multi-dimensional, we fixed the ego velocity, the pedestrian velocity and orientation to visualize the policy for every $s$ and $t$ in the state space. The Frenet frame is relative to the vehicle. A decreasing $s$ means that the pedestrian is closer to the car and $t=\SI{-2}{\meter}$ means the pedestrian is on the right side. We can observe that with a higher velocity, earlier braking is necessary the closer the pedestrian is. Moreover, if the pedestrian is further left, the braking intervention happens later.

\section{Autonomous Emergency Braking System}
The Autonomous Emergency Braking (AEB) system works in combination with the POMDP planner. It uses the generated driving trajectory from the POMDP planner to calculate the risk for a collision. If a collision is unavoidable the AEB system triggers an emergency stop, which has the highest priority and overrules the POMDP planner.

\begin{algorithm}[h]
    \caption{Autonomous Emergency Braking System}\label{euclid}
    \label{alg:pseudo_code_ebs}
    \begin{algorithmic}[1]
        \State \textbf{Input:} Pedestrians position and velocity, ego vehicle trajectory.
        \State Compute TTB given the current ego state.
        \For {each pedestrian}
            \State Predict trajectory given current pedestrian state
            \State Compute the probability of collision $P_c$
            \If { $P_c > P_{c_{\text{threshold}}}$ }
                \State Compute time to collision TTC
                \State Compute risk: $\text{risk} = \min(\frac{\text{TTB}}{\text{TTC}}, 1.0)$ 
                \If { risk $> \text{risk}_{\text{threshold}}$ }
                    \State \textbf{Emergency brake intervention}
                \EndIf
            \EndIf
        \EndFor 
    \end{algorithmic}
\end{algorithm}

The AEB system runs at a higher update rate to detect critical situations as fast as possible, especially when a pedestrian appears behind an occluded area. The system uses, like the POMDP planner, the Frenet frame to generalize the problem to a straight road. The algorithm is described in pseudo code in \cref{alg:pseudo_code_ebs}. The input for the AEB system are the pedestrians position and velocity in the Frenet frame as well as the ego vehicle trajectory given by the higher level planner. 

In the first step, the time-to-brake (TTB) is calculated based on the ego vehicle trajectory. Then a prediction model gives a distribution over possible future states for the pedestrian~\cite{Lefe2014}. This distribution, as well as information on the future ego vehicle state, is used to compute a probability of collision $P_c$. $P_c$ is the estimated  fraction of future pedestrian states overlapping future ego vehicle states. The red circle in \cref{fig:scenerio_general} at $t=\SI{2.5}{\second}$, represents the distribution of possible future states given by the prediction model. The performance of the algorithm is directly related to the quality of the prediction. 

If $P_c$ is above some threshold, we carry an additional check using the following risk metric:
\begin{equation}
    \text{risk} = \min(\frac{\text{TTB}}{\text{TTC}}, 1.0)
\end{equation}
where TTC is the time to collision. If the risk is higher than a defined threshold, an emergency stop is triggered. The implementation of the Autonomous Emergency Braking system is available at~\cite{schr2019}.

\section{Experiments}
\label{sec:Experiments}
We compare three different approaches to get an overview of the advantages and disadvantages of the different systems:
\begin{itemize}
\item Autonomous Emergency Braking
\item POMDP planner
\item POMDP planner with AEB system
\end{itemize}

\begin{table}[t!]
	\centering
	\caption{Simulation parameters}
	\begin{tabular}{lSS}
		\toprule[1pt]
		\text{Parameter} & \text{Value}  \\
        \midrule
        Simulation time step  & \SI{0.05}{\second}  \\
        Pedestrian position tracking standard deviation  & \SI{0.1}{\meter\per\second} \\
        Pedestrian velocity tracking standard deviation  & \SI{0.2}{\meter\per\second} \\
        Object tracking delay  & \SI{200}{\milli\second} \\
        Brake delay  & \SI{200}{\milli\second} \\
        \midrule
        \multicolumn{2}{l}{\textbf{POMDP planner}} \\
		Belief update frequency  & \SI{0.2}{\second}\\
		Decision frequency  & \SI{0.2}{\second} \\
		Pedestrian maximum speed  & \SI{2.0}{\meter\per\second} \\
		\midrule
		\multicolumn{2}{l}{\textbf{Autonomous Emergency Braking System}} \\
		$a_{xmax}$ & \SI{-10.0}{\meter\per\second\squared} \\
		Threshold collision probability & \SI{0.5}{} \\
		Threshold collision risk  & \SI{0.99}{} \\
		\bottomrule[1pt]
	\end{tabular}
	\label{tab:simulation_configuration}
\end{table}

The parameters used for the evaluations are specified in \cref{tab:simulation_configuration}. To evaluate the performance of different implementations, we compare them using scenarios from the EuroNCAP test protocol for vulnerable road users. The aim of the EuroNCAP test protocol is to cover the most amount of accidents. The scenarios in the test protocol are all critical and result in a collision. About 75~\% of all pedestrian accidents are covered with these crossing scenarios \cite{humm2011}. A detailed description of the scenarios can be found at \cite{euro2019}. In the existing version of the test protocol, every scenario has only one defined collision point. To cover a wider variation of collisions along the front of the vehicle (collision grid), we increased the amount of collision points for every defined scenario, see \Cref{tab:euro_ncap_scenarios}.

\begin{table}[t!]
	\centering
	\caption{EuroNCAP VRU scenarios}
	\begin{tabular}{lcccc}
		\toprule[1pt]
		\text{ } & \text{CPAF}  & \text{CPAN-25} & \text{CPAN-75} & \text{CPCN} \\
        \midrule
        Ego velocity [\SI{}{\kilo\meter\per\hour}] & 10-60  & 10-60 & 10-60 & 10-60\\
        Ped velocity [km/h] & 8 & 5 & 5 & 5 \\
        Occlusion  & No & No & No & Yes \\
        Impact point [\%]  & 0--50 & 0--50 & 0--50 & 0--50 \\
		\bottomrule[1pt]
	\end{tabular}
	\label{tab:euro_ncap_scenarios}
\end{table}

\begin{figure}[!t]
\centering
\includegraphics[width=8cm,keepaspectratio=true]{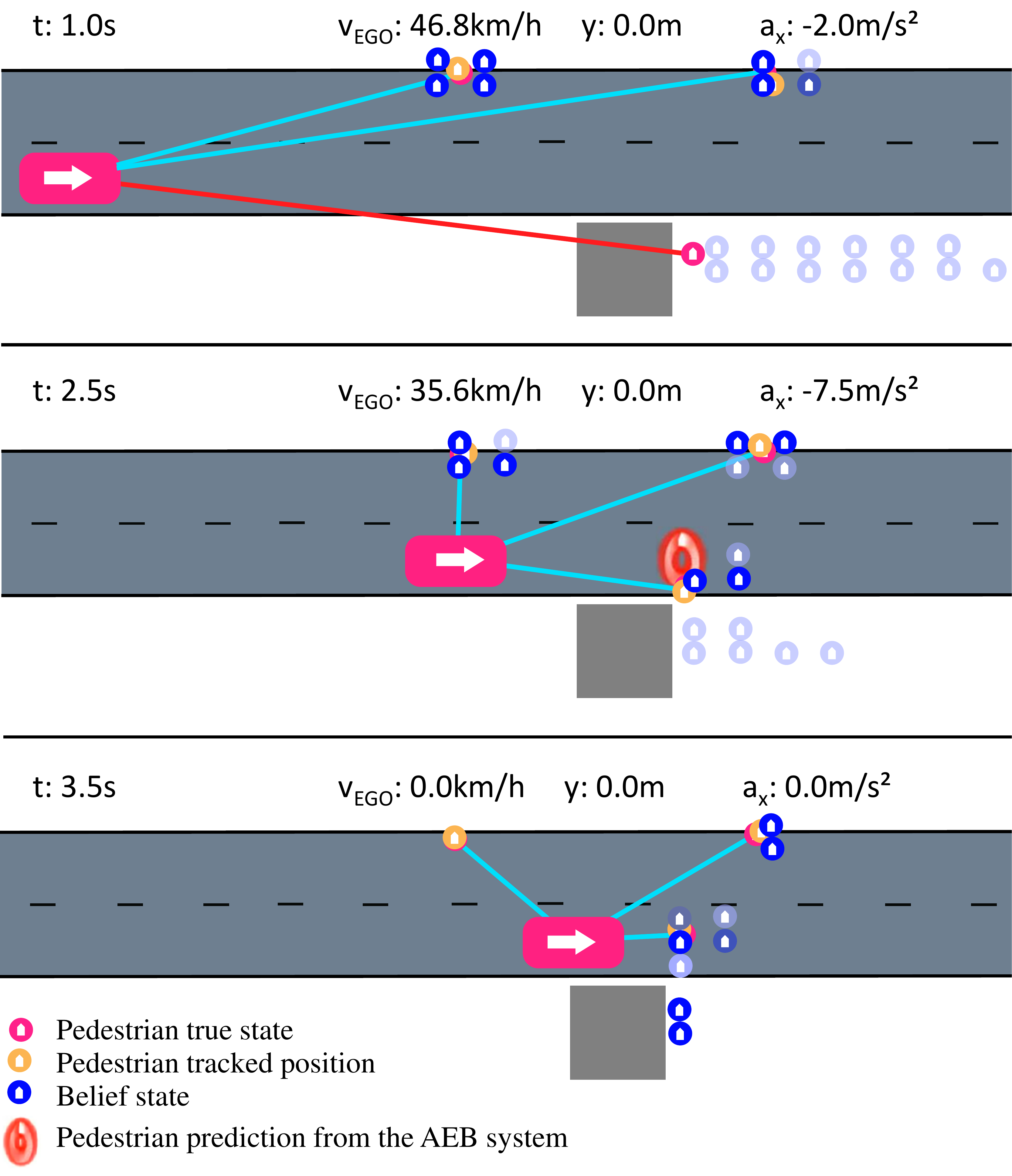}
\caption{Visualization of a scenario with an obstacle on the right side of the road (CPCN) and a crossing pedestrian. In addition two pedestrians stand still on the left side. The belief state is represented in blue. In this case the POMDP planner works in combination with the AEB system. In the top plot, no pedestrian is detected behind the obstacle, but the system maintains an uniform belief over all the possible occluded states. Using this belief the POMDP planner reduces the velocity. In the middle plot, the crossing pedestrian is detected and after performing a belief update, the probability of presence of a pedestrian increases. The red circle represents the predicted pedestrian position from the AEB system, at this time step an emergency braking intervention is triggered to avoid the collision. In the bottom plot, the ego vehicle has stopped and the pedestrian crosses the road.}
\label{fig:scenerio_general} 
\end{figure}

The EuroNCAP test protocol defines different velocities for the ego vehicle. To simplify the analysis, we assume an ego velocity of \SI{50}{\kilo\meter\per\hour}.
Additionally, we added three scenarios to analyze the robustness of the different approaches. In these scenarios, no intervention is required to prevent collision. If the AEB system causes a full brake, it would be a false positive. For two of the scenarios, a pedestrian is crossing the road from the right side; in one scenario the pedestrian is \SI{0.9}{\meter} to the right and in the other \SI{0.9}{\meter} to the left at the passing point. With the third scenario, we evaluate  efficiency in occluded areas. The CPCN scenario (with an obstacle on the right side) is used where no pedestrian is crossing the road. With this scenario we can detect an overly conservative algorithm that would reduce the speed too drastically in the presence of an occluded area. \Cref{fig:scenerio_general} shows examples of the EuroNCAP CPCN scenario with an obstacle on the side where the POMDP planner with the AEB system is active. 

\begin{figure*}[!t]
\centering
\includegraphics[width=16.5cm,keepaspectratio=true]{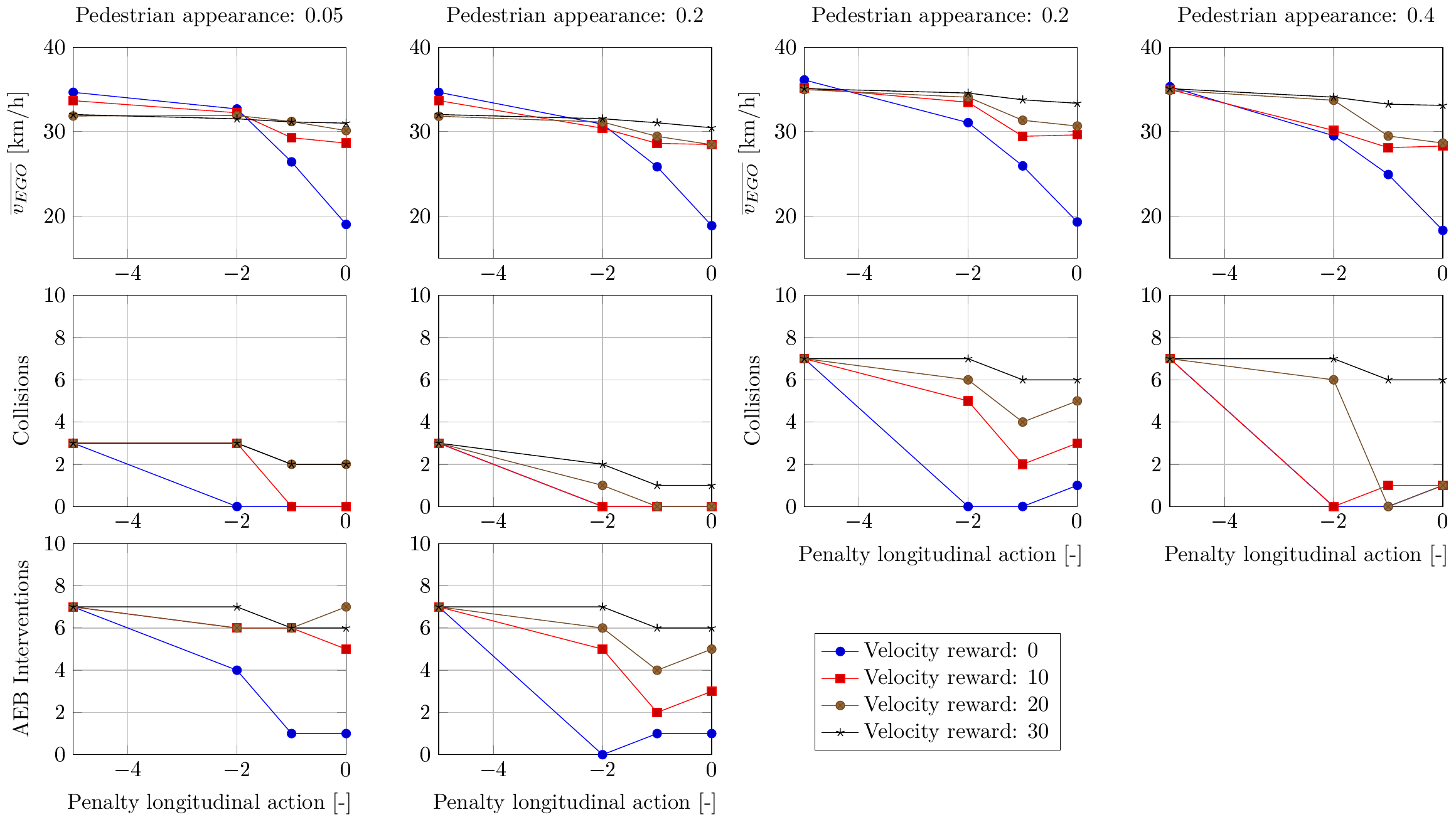}
\caption{The six plots on the left side are the results for the POMDP planner with AEB system and the four diagrams on the right side are the results for the POMDP planner without AEB system. The different lines in each plot represent different velocity rewards. The first row shows the mean velocity versus the probability of a pedestrian appearance and penalty for a longitudinal action. The second row shows the amount of collisions over all scenarios and the third row shows the POMDP planner with AEB system the amount of emergency braking interventions.}
\label{fig:reward_configuration}
\end{figure*}

\subsection{Evaluation metric}

Multiple metrics are available to evaluate performance~\cite{helm2014, chen2017}. We calculate the mean velocity $\overline{\mathbf{v}}$, the mean acceleration $\overline{\mathbf{a}}$, the mean collision velocity $\overline{\mathbf{dv}}$, the number of collisions, and the amount of emergency brake interventions, over all of the EuroNCAP scenarios.

\subsection{Tuning of the reward function}

The behavior of the POMDP planner is influenced by the reward function. The reward parameters must be tuned to fulfil the safety and efficiency requirements. Determining good parameters for the reward function can be challenging. We ran a parameter search and evaluated the resulting policies with the defined scenarios. The objective is to compare the  collision rate, amount of emergency braking, and the mean velocity of the ego vehicle. The following parameters are tuned: 
\begin{itemize}
\item Penalty for longitudinal action (throttle/brake)
\item Velocity reward to keep the velocity close to \SI{50}{\kilo\meter\per\hour}
\item Probability of pedestrian appearance (which is a parameter of our transition model)
\end{itemize}
\Cref{fig:reward_configuration} shows results for different reward functions. We measured the mean velocity, amount of collisions and amount of emergency braking interventions for the resulting policies. The most critical cases are scenarios with an obstacle on the side because reducing the velocity is required  to avoid collision at \SI{50}{\kilo\meter\per\hour}. It is important to notice that the probability of the pedestrian appearance behind an obstacle has a significant influence on the amount of collisions and emergency braking interventions and the mean velocity, which decreases with a higher probability for the pedestrian appearance.

\section{Results}

\begin{figure}[!t]
\centering
\includegraphics[width=7.8cm,keepaspectratio=true]{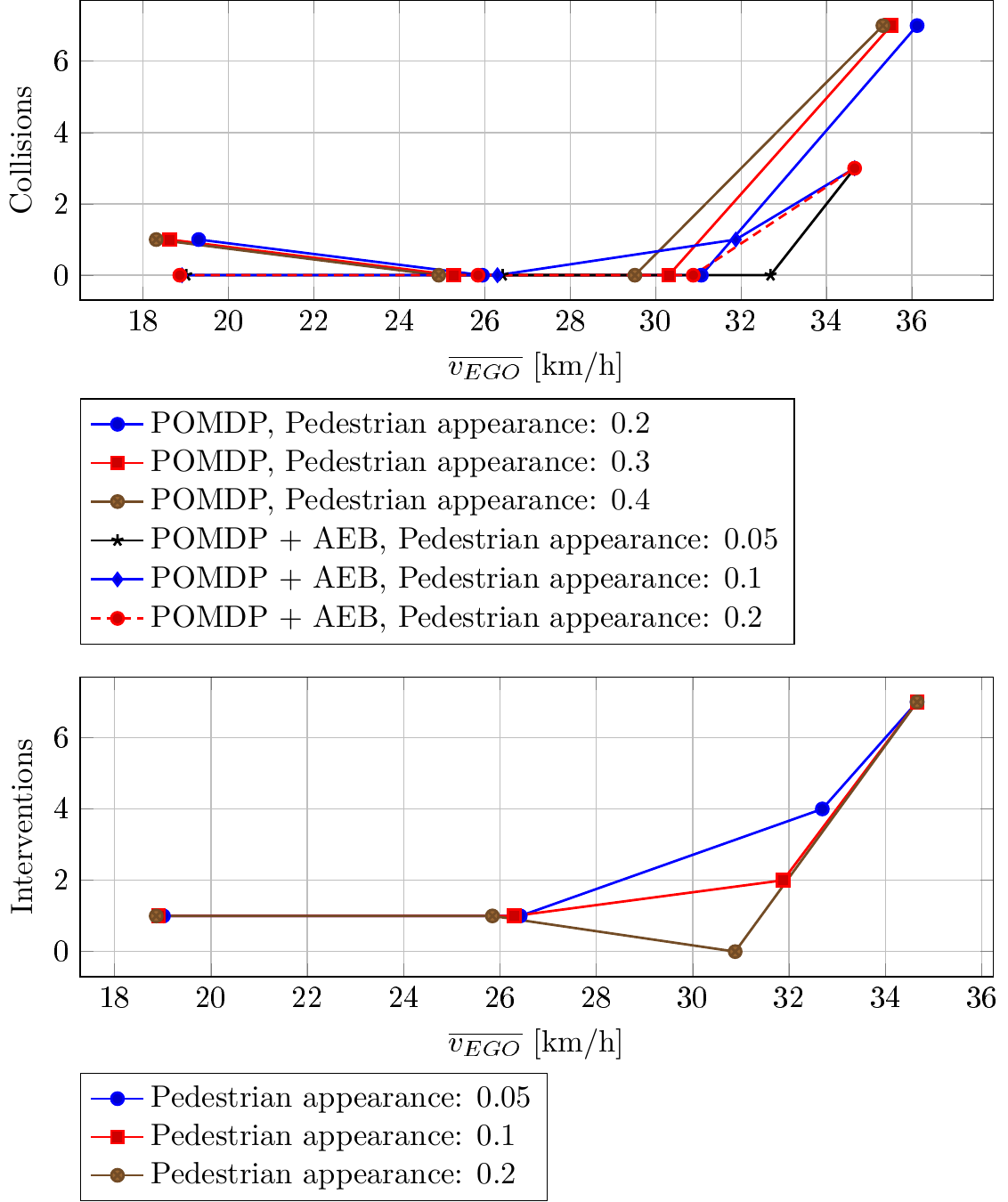}
\caption{The top diagram shows  the amount of collisions for a mean velocity over all tested scenarios for different reward configurations for the POMDP planner with and without AEB system. In the bottom diagram the amount of interventions for the POMDP planner with AEB system is visualised. In both diagrams a higher mean velocity and a lower probability of a pedestrian appearance cause more collisions and more emergency brake interventions.}
\label{fig:simulation_pomdp_configuration} 
\end{figure}

Before comparing the results for the different approaches, we analyze different reward configurations. \Cref{fig:simulation_pomdp_configuration} shows different settings of reward parameters for the POMDP planner with and without the AEB system. In this experiment, the reward for lane keeping and the penalty for a longitudinal action are fixed and the probability of a pedestrian appearance varies. The top plot shows the relation between collisions and mean velocity. A collision-free configuration is possible with both approaches. Combining the POMDP planner with the AEB system results in a higher mean velocity due to the capability of the AEB system to request a stronger brake intervention. The bottom plot shows the number of emergency braking interventions, which decreases when the probability of pedestrian appearance is higher. As the number of interventions decreases, the system behaves more conservatively when passing occluded areas. Based on the results from \Cref{fig:simulation_pomdp_configuration}, we chose the reward parameters that lead to no collisions and the highest mean velocity.

\begin{figure}[!ht]
\centering
\includegraphics[width=8cm,keepaspectratio=true]{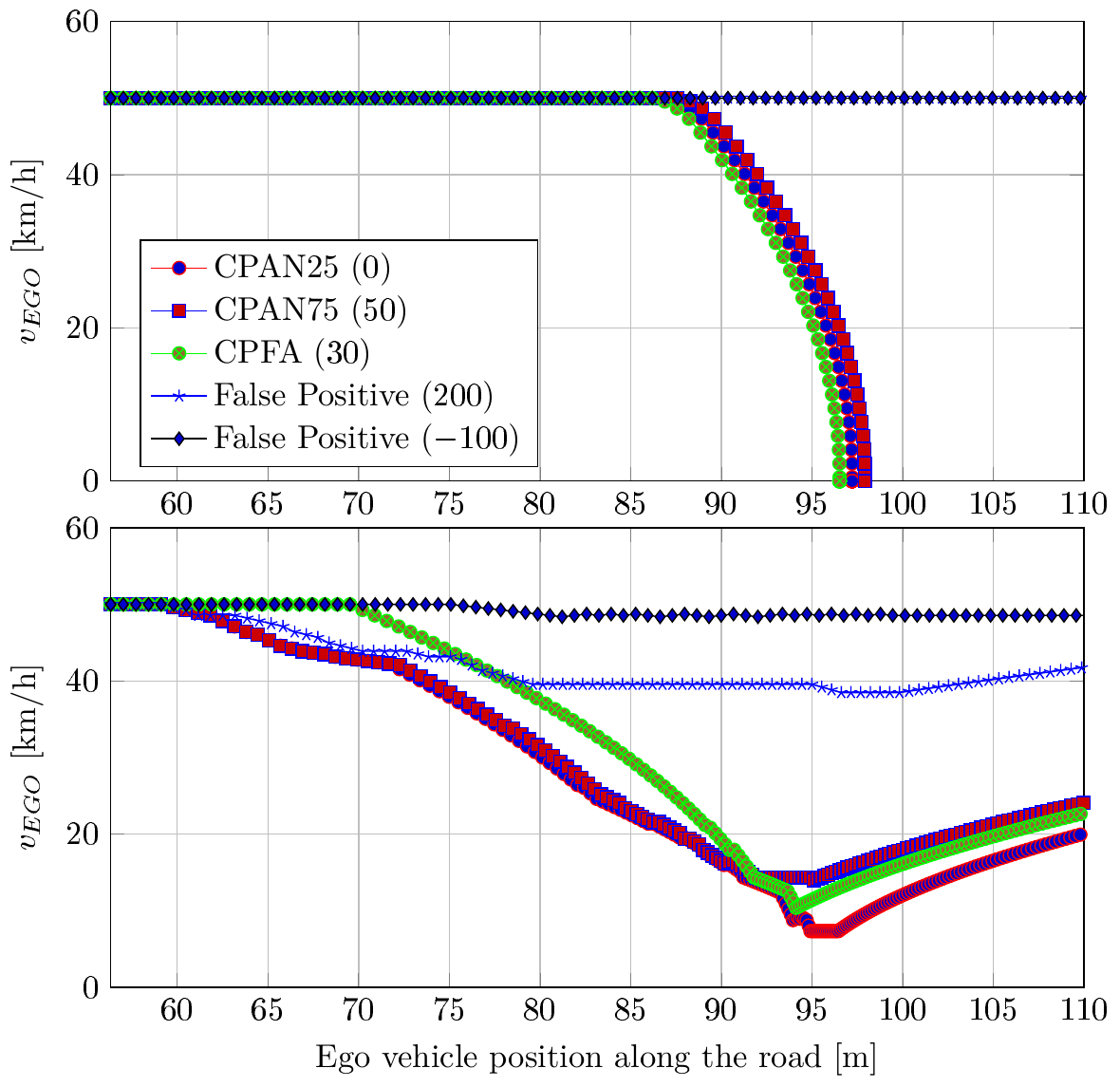}
\caption{Velocity profile for different EuroNCAP scenarios with no occlusions and two false positive test scenarios. Top diagram: The AEB system only brakes at the three collision scenarios. Bottom diagram: The POMDP planner slows down at all scenarios, but with different effects based on the collision point.}
\label{fig:simulation_general_scenarios} 
\end{figure}

\Cref{fig:simulation_general_scenarios} shows the results for the EuroNCAP scenarios without occlusions. The velocity profile of the AEB system is shown on the top, and the velocity profile of the POMDP planner at the bottom. There is no difference between the POMDP planner with and without AEB because there are no occluded areas. The two scenarios, False Positive \num{200} and \num{-100}, are scenarios where the pedestrian is at the passing point of the ego vehicle \SI{0.9}{\meter} to the left and \SI{0.9}{\meter} to the right, respectively. In both cases, the AEB system does not trigger. The POMDP planner behaves differently, reducing the velocity, especially when the pedestrian is directly in front of the vehicle, as shown in scenario False Positive (\num{200}). For all of the three collision scenarios, the POMDP planner slows the vehicle and allows the pedestrian to cross. Afterwards, the ego vehicle accelerates to reach the desired velocity of \SI{50}{\kilo\meter\per\hour}.  

\begin{figure}[!t]
\centering
\includegraphics[width=8cm,keepaspectratio=true]{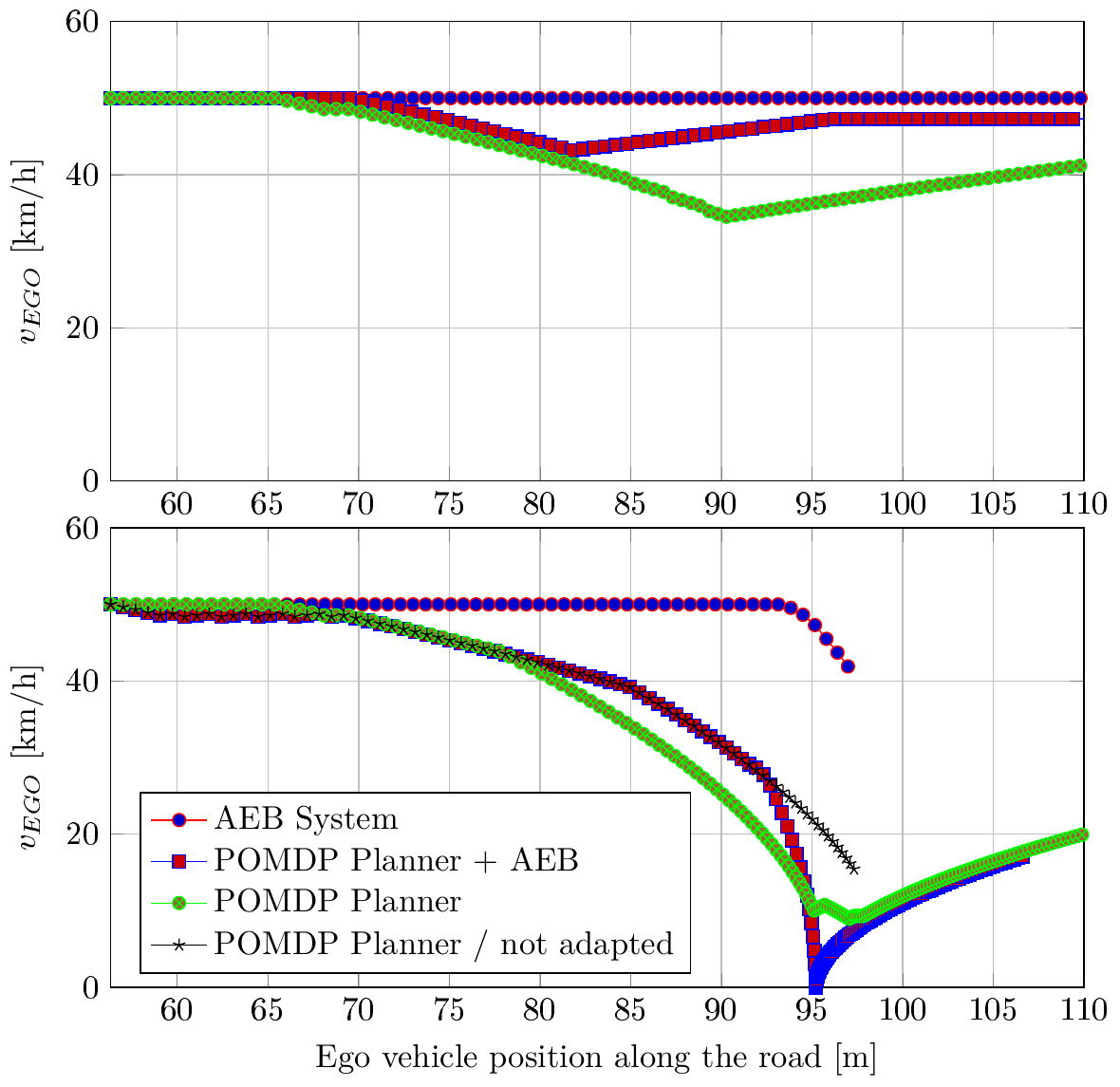}
\caption{Velocity profile for the CPCN scenario with occlusions. Top diagram: No pedestrian crosses the road, but all POMDP planner configurations slow down. Bottom diagram: A pedestrian crosses at 5km/h the road, not all configurations are able to avoid the collision.}
\label{fig:simulation_CPCN_scenario} 
\end{figure}

\Cref{fig:simulation_CPCN_scenario} shows the velocity profiles for the scenario with an occlusion. In the top plot, no pedestrian crosses the road. The AEB system does not decelerate. The two POMDP planners reduce the velocity because of the occluded area. The POMDP planner needs to drive slower than the POMDP planner with the AEB system. The reason for this is illustrated by the bottom figure where a pedestrian crosses the road. The POMDP planner with the AEB system is able to drive faster, but an emergency braking intervention is needed to avoid collision. The POMDP planner decelerates in front of the occluded area. Driving under \SI{50}{\kilo\meter\per\hour} allows it to avoid collision with the occluded pedestrian. In this case, the AEB system is not capable of avoiding collision with a velocity of \SI{50}{\kilo\meter\per\hour}. When driving at high speed, the time to react is not sufficient to stop the vehicle.  \Cref{fig:simulation_CPCN_scenario} shows the velocity profile for a POMDP planner with a deactivated AEB system, which we refer to as not adapted. In this case, the velocity before the obstacle is too high and the deceleration is not sufficient, which illustrates the benefit of the underlying AEB systems.

\Cref{tab:results} summarizes the performance of the three approaches. The AEB system is not able to avoid all collisions, but the two POMDP planners avoid all collisions. The implementation combining the POMDP planner and the AEB system is able to pass obstacles faster. The mean velocity $\overline{\mathbf{v}}$ is higher, but four emergency braking interventions are triggered.

\begin{table}[h!]
	\centering
	\caption{Performance on the different implementations}
	\begin{tabular}{lSSS}
		\toprule[1pt]
		 \text{ } & \text{AEB}  & \text{POMDP}  &  \text{POMDP + AEB} \\
        \midrule
		\text{Collisions} ${[}\#{]}$       & 3  &  0 & 0 \\
		\text{Emergency Brakes} ${[}\#{]}$ & 24 & 0 & 4 \\
		$\overline{\mathbf{dv}} {[}\SI{}{\kilo\meter\per\hour}{]}$  & 3.8 & 0 & 0 \\
        $\overline{\mathbf{v}}$ ${[}\SI{}{\kilo\meter\per\hour}{]}$ &  43.5 & 30.2 & 32.7 \\
        $\overline{\mathbf{a}}$ {[}\SI{}{\meter\per\second\squared}{]} & -8.6 & -3.1 & -3.2 \\
		\bottomrule[1pt]
	\end{tabular}
	\label{tab:results}
\end{table}

\section{Conclusion}
This paper discussed a POMDP approach for a pedestrian collision avoidance system that is capable to handle scenarios with sensor occlusions. The system is able to handle multiple pedestrians while maintaining computational scalability. In addition, an Autonomous Emergency Braking system was implemented to extend the capability in critical situations and increase the driving velocity in non critical situations even in occluded areas.
We used scenarios from the EuroNCAP test protocol for vulnerable road users to evaluate our approach. The experiments showed that different behaviors can be obtained, ranging from a conservative behavior without any emergency brake interventions to a behavior where emergency brakes are always needed to avoid collisions. In the latter case the vehicle passes obstacles on the side of the road at a faster speed.

The investigation showed that defining appropriate parameters for the reward function of the POMDP planner is challenging. The POMDP planner is designed to control the lateral behavior of the vehicle, and it would be interesting to investigate this capability in more depth. 
The implementation of the POMDP planner in combination with the Autonomous Emergency Braking System is publicly available~\cite{scbo2019}.



\section*{ACKNOWLEDGMENT}
\begin{small}
This project received funding from the Electronic Component Systems for European Leadership Joint
Undertaking under grant agreement No 737469. This Joint Undertaking receives support from the European
Union\'s Horizon 2020 research and innovation program and Germany, Austria, Spain, Italy, Latvia, Belgium,
Netherlands, Sweden, Finland, Lithuania, Czech Republic, Romania, Norway. In Austria the project was also funded by the program “IKT der Zukunft” and the Austrian Federal Ministry for Transport, Innovation and Technology (bmvit). The publication was written at VIRTUAL VEHICLE Research Center in Graz and partially funded by the COMET K2 – Competence Centers for Excellent Technologies Programme of the Federal Ministry for Transport, Innovation and Technology (bmvit), the Federal Ministry for Digital, Business and Enterprise (bmdw), the Austrian Research Promotion Agency (FFG), the Province of Styria and the Styrian Business Promotion Agency (SFG).
\end{small}


\printbibliography

\end{document}

%% file: policy/policyplots.tex
\begin{tikzpicture}[]
\begin{groupplot}[
group style={horizontal sep = 0cm, vertical sep = 0.5cm, group size=1 by 3},
width=8.5cm,
height=3cm
]
\nextgroupplot [
ylabel = {t [m]}, 
title style={at={(0.73,0.85)},anchor=north},
title = {Ego velocity: 14.4 km/h}, 
xmin = {0}, xmax = {50}, ymax = {5}, 
ymin = {-5}, enlargelimits = false, axis on top, 
colormap={mycolormap}{ rgb(0cm)=(0.001462,0.000466,0.013866) rgb(1cm)=(0.016561,0.013136,0.080282) rgb(2cm)=(0.051644,0.032474,0.159254) rgb(3cm)=(0.09299,0.045583,0.234358) rgb(4cm)=(0.149073,0.045468,0.317085) rgb(5cm)=(0.211095,0.03703,0.378563) rgb(6cm)=(0.271347,0.040922,0.411976) rgb(7cm)=(0.328921,0.057827,0.427511) rgb(8cm)=(0.379001,0.076253,0.432719) rgb(9cm)=(0.434987,0.097069,0.432039) rgb(10cm)=(0.491022,0.117179,0.425552) rgb(11cm)=(0.547157,0.136929,0.413511) rgb(12cm)=(0.603139,0.157151,0.395891) rgb(13cm)=(0.652369,0.176421,0.375586) rgb(14cm)=(0.7065,0.200728,0.347777) rgb(15cm)=(0.758422,0.229097,0.315266) rgb(16cm)=(0.807082,0.262692,0.278898) rgb(17cm)=(0.846709,0.297559,0.244113) rgb(18cm)=(0.886302,0.342586,0.202968) rgb(19cm)=(0.919879,0.393389,0.16007) rgb(20cm)=(0.946965,0.449191,0.115272) rgb(21cm)=(0.967322,0.509078,0.068659) rgb(22cm)=(0.979666,0.565057,0.031409) rgb(23cm)=(0.986964,0.630485,0.030908) rgb(24cm)=(0.987124,0.697944,0.087731) rgb(25cm)=(0.980032,0.766837,0.166353) rgb(26cm)=(0.966243,0.836191,0.261534) rgb(27cm)=(0.951546,0.896226,0.365627) rgb(28cm)=(0.949545,0.955063,0.50786) rgb(29cm)=(0.988362,0.998364,0.644924) }
]

\addplot [point meta min=-4.0, point meta max=1.0] graphics [xmin=0, xmax=50, ymin=-5, ymax=5] {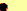};
\nextgroupplot [
ylabel = {t [m]},
title style={at={(0.73,0.85)},anchor=north},
title = {Ego velocity: 27.0 km/h}, xmin = {0}, xmax = {50}, ymax = {5}, ymin = {-5}, enlargelimits = false, axis on top,
colormap={mycolormap}{ rgb(0cm)=(0.001462,0.000466,0.013866) rgb(1cm)=(0.016561,0.013136,0.080282) rgb(2cm)=(0.051644,0.032474,0.159254) rgb(3cm)=(0.09299,0.045583,0.234358) rgb(4cm)=(0.149073,0.045468,0.317085) rgb(5cm)=(0.211095,0.03703,0.378563) rgb(6cm)=(0.271347,0.040922,0.411976) rgb(7cm)=(0.328921,0.057827,0.427511) rgb(8cm)=(0.379001,0.076253,0.432719) rgb(9cm)=(0.434987,0.097069,0.432039) rgb(10cm)=(0.491022,0.117179,0.425552) rgb(11cm)=(0.547157,0.136929,0.413511) rgb(12cm)=(0.603139,0.157151,0.395891) rgb(13cm)=(0.652369,0.176421,0.375586) rgb(14cm)=(0.7065,0.200728,0.347777) rgb(15cm)=(0.758422,0.229097,0.315266) rgb(16cm)=(0.807082,0.262692,0.278898) rgb(17cm)=(0.846709,0.297559,0.244113) rgb(18cm)=(0.886302,0.342586,0.202968) rgb(19cm)=(0.919879,0.393389,0.16007) rgb(20cm)=(0.946965,0.449191,0.115272) rgb(21cm)=(0.967322,0.509078,0.068659) rgb(22cm)=(0.979666,0.565057,0.031409) rgb(23cm)=(0.986964,0.630485,0.030908) rgb(24cm)=(0.987124,0.697944,0.087731) rgb(25cm)=(0.980032,0.766837,0.166353) rgb(26cm)=(0.966243,0.836191,0.261534) rgb(27cm)=(0.951546,0.896226,0.365627) rgb(28cm)=(0.949545,0.955063,0.50786) rgb(29cm)=(0.988362,0.998364,0.644924) }
]
\addplot [point meta min=-4.0, point meta max=1.0] graphics [xmin=0, xmax=50, ymin=-5, ymax=5] {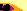};
\nextgroupplot [
ylabel = {t [m]}, 
title style={at={(0.73,0.85)},anchor=north},
title = {Ego velocity: 46.8 km/h}, xmin = {0}, xmax = {50}, ymax = {5}, xlabel = {s [m]}, ymin = {-5},
enlargelimits = false, axis on top,
colorbar,
colorbar horizontal,
    colorbar style={
        height=0.5cm,
        ylabel={\SI{}{\meter\per\second\squared}},
        ytick={-4 , 2},
        at={(0.0, -0.8) anchor=north west}
    },
colormap={mycolormap}{ rgb(0cm)=(0.001462,0.000466,0.013866) rgb(1cm)=(0.016561,0.013136,0.080282) rgb(2cm)=(0.051644,0.032474,0.159254) rgb(3cm)=(0.09299,0.045583,0.234358) rgb(4cm)=(0.149073,0.045468,0.317085) rgb(5cm)=(0.211095,0.03703,0.378563) rgb(6cm)=(0.271347,0.040922,0.411976) rgb(7cm)=(0.328921,0.057827,0.427511) rgb(8cm)=(0.379001,0.076253,0.432719) rgb(9cm)=(0.434987,0.097069,0.432039) rgb(10cm)=(0.491022,0.117179,0.425552) rgb(11cm)=(0.547157,0.136929,0.413511) rgb(12cm)=(0.603139,0.157151,0.395891) rgb(13cm)=(0.652369,0.176421,0.375586) rgb(14cm)=(0.7065,0.200728,0.347777) rgb(15cm)=(0.758422,0.229097,0.315266) rgb(16cm)=(0.807082,0.262692,0.278898) rgb(17cm)=(0.846709,0.297559,0.244113) rgb(18cm)=(0.886302,0.342586,0.202968) rgb(19cm)=(0.919879,0.393389,0.16007) rgb(20cm)=(0.946965,0.449191,0.115272) rgb(21cm)=(0.967322,0.509078,0.068659) rgb(22cm)=(0.979666,0.565057,0.031409) rgb(23cm)=(0.986964,0.630485,0.030908) rgb(24cm)=(0.987124,0.697944,0.087731) rgb(25cm)=(0.980032,0.766837,0.166353) rgb(26cm)=(0.966243,0.836191,0.261534) rgb(27cm)=(0.951546,0.896226,0.365627) rgb(28cm)=(0.949545,0.955063,0.50786) rgb(29cm)=(0.988362,0.998364,0.644924) }
]
\addplot [point meta min=-4.0, point meta max=1.0] graphics [xmin=0, xmax=50, ymin=-5, ymax=5] {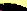};
\end{groupplot}

\end{tikzpicture}